\documentclass[10pt,twocolumn,letterpaper]{article}

\usepackage{iccv}
\usepackage{times}
\usepackage{epsfig}
\usepackage{graphicx}
\usepackage{amsmath}
\usepackage{amssymb}
\usepackage[export]{adjustbox}
\usepackage{booktabs}
\usepackage{multirow}
\usepackage{algorithm}  
\usepackage{algorithmicx}  
\usepackage{algpseudocode}  
\usepackage{cite}
\usepackage{subfig}
\usepackage{array}
\usepackage{colortbl}
\usepackage[accsupp]{axessibility}
\usepackage{url}

\usepackage[table]{xcolor}
\definecolor{col1}{RGB}{232, 161, 148}
\definecolor{col2}{RGB}{148, 187, 232}

\usepackage[pagebackref=true,breaklinks=true,letterpaper=true,colorlinks,bookmarks=false]{hyperref}

\iccvfinalcopy 

\def\iccvPaperID{4076} 

\ificcvfinal\fi
\begin{document}

\title{Single-Camera 3D Head Fitting for Mixed Reality Clinical Applications}

\author{Tejas Mane\textsuperscript{1}, Aylar Bayramova\textsuperscript{1}, Kostas Daniilidis\textsuperscript{2}, Philippos Mordohai\textsuperscript{3}, Elena Bernardis\textsuperscript{1},\\
\textsuperscript{1}Dept. Dermatology, University of Pennsylvania, Philadelphia, PA 19104, USA\\
\textsuperscript{2}Dept. Computer and Information Science, University of Pennsylvania,  Philadelphia, PA 19104, USA\\
\textsuperscript{3}Dept. Computer Science, Stevens Institute of Technology, Hoboken, NJ, 07030, USA 

}
\twocolumn[{
\vspace{-1em}
\maketitle
\vspace{-1em}
}]
\ificcvfinal\thispagestyle{empty}\fi

\begin{abstract}
   We address the problem of estimating the shape of a person's head, defined as the geometry of the complete head surface, from a video taken with a single moving camera, and determining the alignment of the fitted 3D head for all video frames, irrespective of the person's pose. 3D head reconstructions commonly tend to focus on perfecting the face reconstruction, leaving the scalp to a statistical approximation. Our goal is to reconstruct the head model of each person to enable future mixed reality applications. To do this, we recover a dense 3D reconstruction and camera information via structure-from-motion and multi-view stereo. These are then used in a new two-stage fitting process to recover the 3D head shape by iteratively fitting a 3D morphable model of the head with the dense reconstruction in canonical space and fitting it to each person's head, using both traditional facial landmarks and scalp features extracted from the head's segmentation mask. Our approach recovers consistent geometry for varying head shapes, from videos taken by different people, with different smartphones, and in a variety of environments from living rooms to outdoor spaces.
\end{abstract}

\let\thefootnote\relax\footnotetext{This manuscript has been accepted for publication at Computer Vision and Image Understanding. DOI: 10.1016/j.cviu.2022.103384}
\vspace{-10pt}
\section{Introduction}
3D head and 3D hair reconstruction models have been popular topics in computer vision over the past years as their usefulness spans a broad range of applications. 
Current 3D reconstructions  \cite{guo2020towards, tran2018nonlinear,tuan2017regressing,zhu2017face, cao2018stabilized, kim2018deep,Alldieck2018,Beeler2010,Agrawal_2020_WACV} heavily rely on facial landmarks, usually leaving the back of the head unaccounted for. Thus, the resulting head reconstructions tend to be very precise in terms of facial features but include only a statistical approximation of the rest of the head. 3D hair reconstruction \cite{Zhou_2018_ECCV,3DHair_from_video_Ira}, which often also relies on these `face-centric' 3D head fitting models, additionally focuses on rendering a photo-realistic appearance of the hair, and not on representing the actual clinical condition of the person's hair. 
For clinical applications, the reconstructed shape has to resemble the complete head shape of the person as close as possible. 
Furthermore, to map information over the entire scalp, such as hair or skin variations, to track changes over time, we need to recover the alignment of the head for the entire pose range, including top and back views where the face is occluded and traditional facial landmark detectors cannot be applied.  
This has been achieved with high-end multi-camera setups \cite{alexander2013digital,Cao2018,fyffe_multi-view_2017} or 3D scanners \cite{debevec_facialScans,lee1995realistic}. However, reconstructing accurate head shape requires setups that are not easily scalable to the clinical practice nor to the general population. 

\begin{figure*}
\centering
\includegraphics[width=0.98\linewidth]{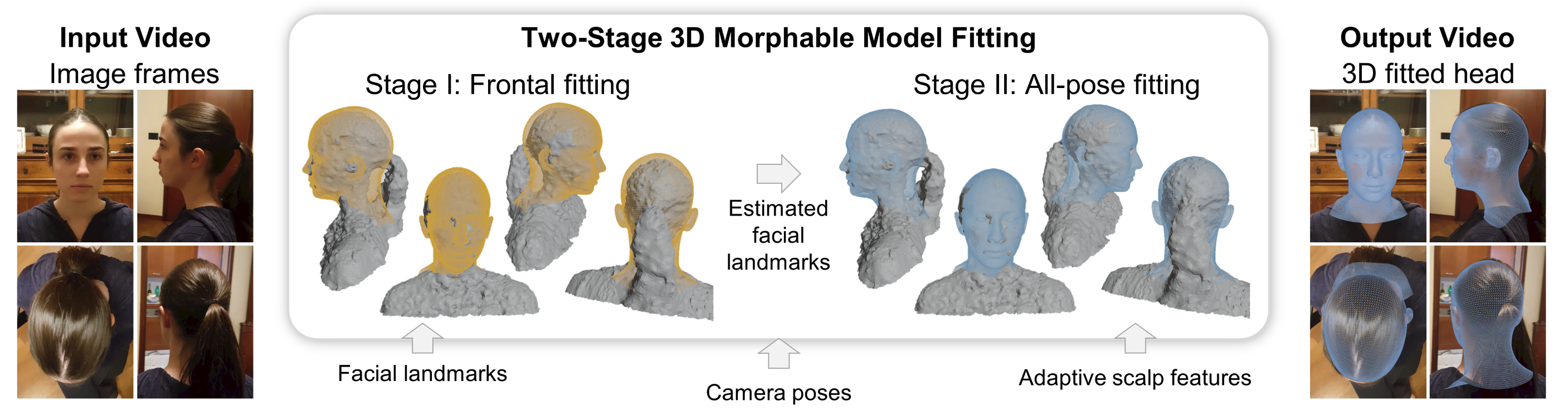} 
\caption{\textbf{All-pose 3D head fitting overview.} Given a video of a person's head, we detect facial landmarks on frontal frames and run a photogrammetry pipeline to recover a generic 3D dense reconstruction $\mathcal{G}$ (gray models) and camera poses. We then use silhouettes extracted from $\mathcal{G}$ to define adaptive scalp features to better encode the shape of the scalp. All extracted keypoints, $\mathcal{G}$, and cameras are then input into a two-stage fitting algorithm: Stage 1 fits the 3DMM for frontal views using facial landmarks to yield updated estimated facial landmarks for images sampled from the complete pose range; Stage 2 uses these sampled images, estimated facial landmarks together with the scalp features, for all-pose fitting. 
}
\label{fig:overview}  
\end{figure*}

To create 3D models of the head that may be used for future augmented or mixed reality applications, we present a 3D morphable model (3DMM) head fitting framework that only requires a video taken with a smartphone or off-the-shelf digital camera, to recover the surface geometry of the person's head. 3DMMs offer a consistent 3D mesh topology that will easily allow, in the future, to compare a specific location on the person's scalp at different time-points as well as between subjects. 
Solving this problem for any hair volume scenario is very challenging as it requires estimating the shape of the scalp even when it is entirely occluded. The main focus of this work is on reconstructing the 3D geometry of subjects whose scalp outline is either partially visible due to areas of low hair density or contoured closely by hair with minimal volume. 
Given a video of a person's head taken with a single moving camera, we fit a 3DMM of the complete head in a two-stage fitting approach using traditional facial landmarks as well as adaptive scalp features. 
As outlined in Figure \ref{fig:overview}, to recover the head geometry, we fit a 3DMM of the head via a two-stage fitting algorithm. Both fitting steps are done via an iterative optimization that uses the SfM 3D head reconstruction and camera poses to refine the 3DMM with respect to the camera poses. Stage 1 fits the frontal views using facial landmarks to yield estimated facial landmarks for all frames/camera poses; Stage 2 uses images sampled from the complete pose range, estimated facial landmarks together with the scalp features for all-pose fitting. 
The output is a 3D head geometry that captures the surface of the complete head and closely contours the scalp region. This model can then be anchored over the entire pose range for future mixed reality applications.
Our contributions can be summarized as follows.  
We propose:
i) a two-stage 3D head fitting algorithm that incorporates face and scalp information by including also new adaptive scalp features to encode the shape of the entire head; 
ii) a solution to enable 3DMM fitting across the complete pose range that combines traditional 3DMM fitting with SfM to get the best of both worlds: accurate camera poses and reconstruction.
As shown in the results section, the approach works on a variety of videos taken by different people, with different smartphones and in a variety of static environments from living rooms to outdoor spaces. 

\section{Related Work}
In recent years, there has been significant progress in recovering 3D shapes from 2D images \cite{goel2020shape} to recover faces \cite{blanz1999morphable,SMPL:2015} as well as animals and objects like birds, cars, zebras, airplanes, \etc \cite{badger20203d,goel2020shape,kanazawa2018learning, zuffi2019three}. A common approach is to use 3DMMs to represent the 3D shape as an average shape plus a per-instance predicted deformation. 
There are several popular 3DMM models for the human head \cite{huber2016multiresolution, bfm09}. However, few models exist for the complete head (face and scalp) \cite{cao2013facewarehouse,Dai2019,FLAME:SiggraphAsia2017,ploumpis2020towards}. The Liverpool Head  Model (LHM) \cite{Dai2019}, accounting for head shape information from over 1200 subjects, was among the first to focus on modeling also the accurate shape of the scalp portion of the head. Its successor, the Universal Head Model (UHM) \cite{ploumpis2020towards}, also added more facial details via the existing Large Scale Facial Model \cite{Booth2018_IJCV}. 
3DMMs have also been utilized to develop deep learning models for fitting 3D face/head models from individual images \cite{cao2018stabilized, guo2020towards, kim2018deep, tran2018nonlinear,tuan2017regressing,zhu2017face}. However, they are usually limited to pose ranges dictated by state-of-the-art facial landmark detectors. The fitting problem in images where faces cannot be detected has yet to be addressed. 

3D hair reconstruction, from strand-accurate geometric reconstruction of the individual hair \cite{Nam2019_hair3D} to overall hair volume reconstruction from single images \cite{Zhou_2018_ECCV} or videos \cite{3DHair_from_video_Ira} also relies on underlying head models and hair masks segmentation. \cite{Zhou_2018_ECCV} 
present a single-view hair reconstruction by fitting a 3DMM similarly to \cite{hu2017avatar}. To estimate head shape, \cite{3DHair_from_video_Ira} fuse the Basel face model \cite{blanz1999morphable} with the generic full head model \cite{cao2013facewarehouse} to get the complete 3D head. Again, the retrieval of the 3D head is based on facial landmark fitting alone and the back of the head is usually left unaccounted for. Additionally, all of these models tend to focus on outlining the hair \vs rest, regardless of scalp boundaries or variations in hair density.

Structure-from-Motion (SfM) \cite{Hartley:2003:MVG:861369} with Multi-View-Stereo (MVS) \cite{furukawa15} aim to reconstruct a 3D model of a scene using a set of images taken from different viewpoints. Meshroom \cite{Meshroom}, for example, is a popular 3D photogrammetry-based software that combines SfM and MVS \cite{Jancosek2011,Moulon2012} to reconstruct a 3D scene and recover estimated camera position for each frame. Another alternative for open source SfM+MVS package is COLMAP \cite{schonberger2016structure, schoenberger2016mvs}. 
While usually requiring a static environment, they allow for the computation of correspondences between images with high accuracy exploiting camera motion, so they are robust to input variations and can handle arbitrary backgrounds and large pose variations.  

\section{Approach}

Given a video of a person's head taken with a single moving camera, we want to recover the head geometry (face and scalp without hair volume).   
As outlined in Figure \ref{fig:overview}, we start by extracting initial facial landmarks for each applicable input frame. In parallel, we apply an available photogrammetry pipeline, leveraging SfM and MVS, to get a generic 3D dense reconstruction $\mathcal{G}$ of the scene while recovering camera information for every frame. The 3D reconstruction is used to obtain the silhouettes (the projection of the dense reconstruction on each frame), which in turn can be used to extract scalp features for any image. 
Extracted keypoints, together with $\mathcal{G}$ and cameras are then inputted into a two-stage fitting algorithm: Stage 1 fits the 3DMM for frontal views using facial landmarks to yield estimated facial landmarks; Stage 2 then uses these sampled images, estimated facial landmarks together with the scalp features and the camera poses, for all-pose fitting. 

\subsection{Pre-processing} 
\label{sec:preprocess} 
We extract $N$ uniformly sampled frames $\{I_n\}_{n=1}^{N}$ from each video. We start by detecting facial landmarks and applying a SfM and MVS pipeline to get a 3D dense reconstruction of the person's head and recover camera information for every frame. 

\textbf{3D generic reconstruction.} 
We use Meshroom \cite{Meshroom} to create a 3D dense reconstruction of the scene and recover perspective camera matrices for each frame, up to scale. Since we take close-up videos of people, mesh structures of the background will be comprised of large triangles, and they may include several disconnected components. For all videos, the largest connected component always includes the person's head, neck and torso, so we discard the remaining smaller components. We used mesh filtering to remove larger triangles to obtain a generic 3D reconstruction $\mathcal{G}$ of the head, neck and upper torso regions of the person. 
We then extract the corresponding silhouette $\mathcal{G}_n$ by projecting the mesh onto each frame $I_n$.  

\textbf{Initial facial landmarks.} To extract facial landmarks, we start by separating out `frontal' frames (covering approximately the frontal half of the head, thus including also profile ones) in which a face can be detected using the Dlib library \cite{dlib09}.  We then apply an open source 3D facial landmark detector \cite{bulat2017far} to determine landmarks on the nose, eyes, mouth, and jawlines. This allows us to obtain a small set of images $N_f$ to encode the geometry of the head using facial landmarks.

\subsection{Universal Head Model} 
\label{sec:3Dhead} 
To recover the head geometry, we fit a 3DMM of the head. Since we are interested in fitting the complete head (face and scalp), we chose the \textit{Universal Head Model} (UHM) \cite{ploumpis2020towards}. The UHM model is a PCA-based morphable head model where each head $S$, defined as the $\mathbb{R}^3$ position vector of its $V$ vertices, can be represented as the average head shape $S^{mean}$ plus a per instance deformation:
\begin{align}\label{Eqn:UHMshapeDef}
    S(\alpha) = S^{mean} +  U\alpha,
\end{align}
where $\alpha$ is the $n_{\alpha} \times 1$ shape parameters vector  associated with the $3V \times n_\alpha$ principal components $U$ of the UHM model with $n_\alpha$ the number of shape parameters. 

\setlength{\tabcolsep}{0pt}
\begin{figure}[t!]
\centering
\begin{tabular}{c@{\hskip 0.3in}c@{\hskip 0.3in}c@{\hskip 0.3in}r}
\textbf{a:} $y_{k,n}$, $\mathcal{G}_n$ & \textbf{b:} $\mathcal{G}$ & \textbf{c:} $S(\alpha)$  &  \textbf{d:} $T_{sim}(S(\alpha))$ \\ 
\end{tabular}
\includegraphics[width=0.98\linewidth]{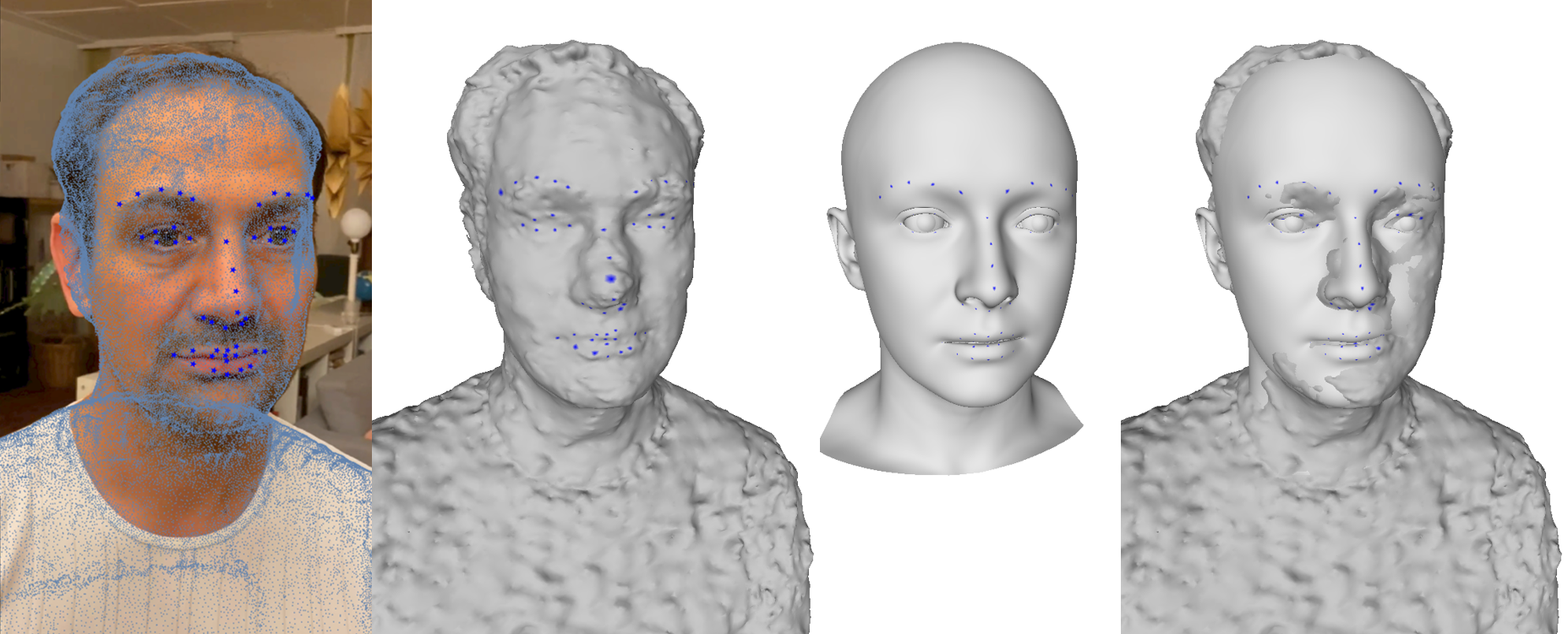} %
\caption{\textbf{Sample alignment of the UHM model based on frontal image with dense reconstruction $\mathcal{G}$ prior to nonlinear optimization.} We show: a) a frontal image with projected $\mathcal{G}$ and detected facial landmarks; b) $\mathcal{G}$ with the determined facial landmarks (blue); c) the UHM model with the facial landmarks (blue); d) the UHM model aligned with the $\mathcal{G}$ using $T_{sim}$. }
\label{fig:align}
\end{figure}

\subsection{Two Stage Fitting Algorithm} 
\label{sec:2stageFitting} 
\textbf{Frontal fitting using facial landmarks (Stage 1).} Using a frontal frame, we first determine the facial landmark vertices on the SfM mesh $\mathcal{G}$ and use them to obtain $T_{sim}(S(\alpha^{i}))$, 
the aligned UHM mesh $S(\alpha^i)$ with $\mathcal{G}$, where $S(\alpha^{i})$ is initially set to the mean model $\mu$ for $i=0$ and $T_{sim}$ is the similarity transformation 
between $\mathcal{G}$ and $S(\alpha^{i})$ obtained via 
\cite{umeyama1991least} as shown in Figure \ref{fig:align}. 
Once the alignment is known, we can then use the perspective cameras estimated during SfM to obtain the projection of the UHM mesh for any image $I_{n}$. Due to the jagged nature of the SfM mesh, however, we first have to ensure that $T_{sim}$ is the optimal rigid transformation in the SfM space with respect to the estimated perspective cameras. 

Let $X_k$ be the 3D coordinates of the keypoints of the UHM model ($S(\alpha^i)$) and $\pi_n$ be the camera projection function for image $I_n$.
To find the optimized similarity matrix $T_{opt}$ for a given $\pi_{n}$, 
we minimize the error between keypoints $y_{k,n}$ on the images and the corresponding image coordinates of the UHM keypoints, 
\begin{align}
\sum_n\sum_k
||\pi_{n}(T_{opt} T_{sim} X_k)  - y_{k,n}||^2,
\label{eq:initAlign}
\end{align}
using the Levenberg-Marquardt optimization algorithm \cite{more1978levenberg}. The UHM model $T_{sim}(S(\alpha_{i}))$ updated at every iteration via the Umeyama algorithm provides a good initial solution for the 
optimization algorithm.

\begin{figure}
\centering
\includegraphics[width=0.99\linewidth]{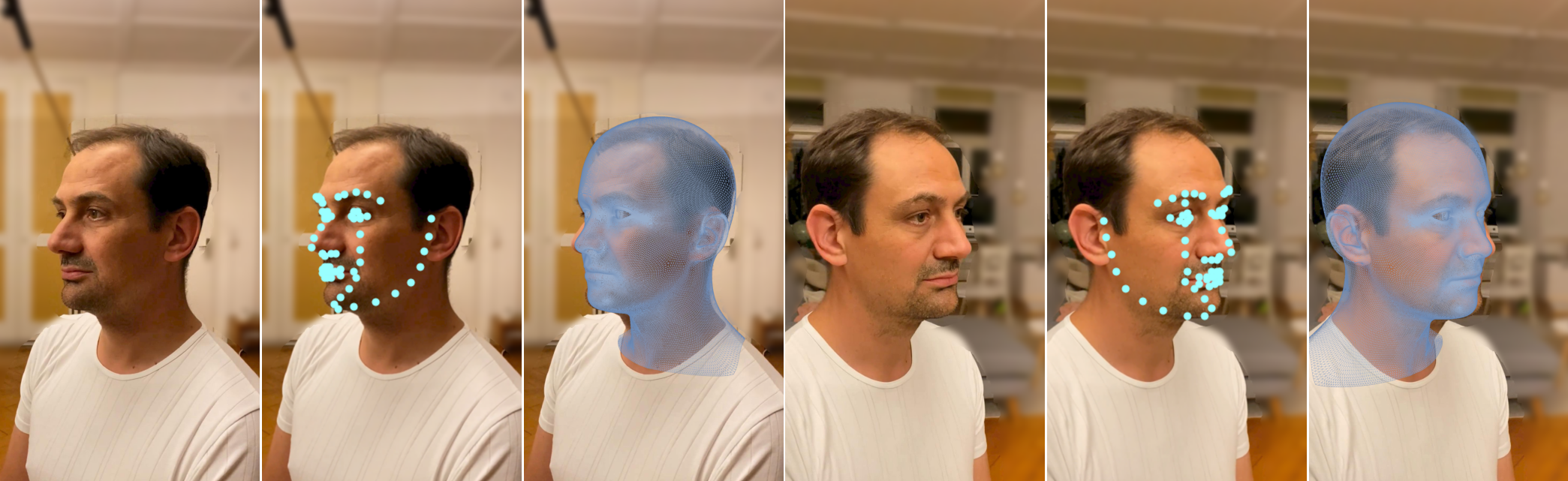} \\
\caption{ \textbf{Frontal fitting.} For two sample views, we show: input image $I_n$; initial facial landmarks;  
and frontal fit $S_n^{front}$, resulting from Stage 1, projected back on each frame.}
\label{fig:frontal}
\end{figure}

Let $\mathcal{T}_{S,n}= \Pi_{n}T_{opt}T_{sim}$ be the resulting transformation between the UHM canonical coordinate system and the coordinate system of image $I_n$, where $\Pi_{n}$ is the camera projection matrix. Note that the image coordinates obtained by $\pi_{n}(T_{opt}T_{sim}X_{k})$ are the same as the coordinates of $\mathcal{T}_{S,n}X_{k}$ divided by the depth coordinate (z axis) of $\mathcal{T}_{S,n}X_{k}$. We can then solve for the optimal UHM fit by extending the algorithm described by \cite{hu2017efficient} to multiple images and minimizing the distance between 3D points on the UHM mesh and the back-projected keypoints $Y_{k,n}$
detected in the images, 
\begin{gather}\label{eq:multiImageFitting}
\sum_n \sum_k ||\mathcal{T}_{S,n}X_k - 
Y_{k,n}
||^2 + \lambda (\alpha^i)^T \sigma_{v}^{-1} \alpha^i,  
\end{gather}
where $\sigma_{v}$ is the diagonal matrix of the eigenvalues of the UHM model, $\lambda$ is a regularization term to prevent extreme unnatural shapes, and $Y_{k,n} = \pi_{n}^{-1}(y_{k,n},z_{k,n})$ the back-projected keypoints detected in the images (recall that the camera poses required for projection are obtained from the SfM step). $\pi_{n}^{-1}$ is the backprojection function that reconstructs a 3D point given pixel coordinates $y_{k,n}$ into the camera coordinate system for $I_n$ and the corresponding depth $z_{k,n}$ estimated from the depth coordinate of the corresponding landmark of the UHM model.
Next, let $\mathcal{H} = U \mathcal{T}_{S,n}$ be the
$3K\times n_\alpha$ matrix of the transformed principal components of the UHM model.
The equation can now be linearized to solve for the optimal parameters $\alpha$:
\begin{gather}
   \alpha^{i+1} = \left( \sum_n \mathcal{H}^{T} \mathcal{H} + \lambda  \sigma_{v}^{-1} \right)^{-1}  
   \sum_n 
   \mathcal{H}^{T} (
   \widehat{Y_n} - 
   \widehat{\mathcal{T}_{S,n} \mu}), 
   \label{eq:multiImageFittingLin}
\end{gather}
where $K$ is the number of keypoints (a subset of V), $\widehat{}$ denotes the corresponding stacked vector of dimension $3K \times 1$, and $\sigma_{v}^{-1}$ is a $n_\alpha \times n_\alpha$ diagonal matrix with the eigenvalues.
We iterate this process to obtain frontal shape parameters $\alpha_{front}$. 

Figure \ref{fig:frontal} shows sample output of Stage 1 based only on frontal fitting. While a good starting point, it fails to capture the shape of the scalp for the subject. We thus need additional features
to add constraints on the fitted mesh so that it passes through these points and thus close to the scalp.
Note that we can now use $S^{front} = S(\alpha_{front})$ to estimate landmark positions on all images regardless of the pose. Hence, we repeat the fitting procedure for the complete pose range using the estimates of the landmark positions from $S^{front}$ and additional scalp features as described next. 

\begin{figure}[t]
\centering
\includegraphics[width=0.95\linewidth]{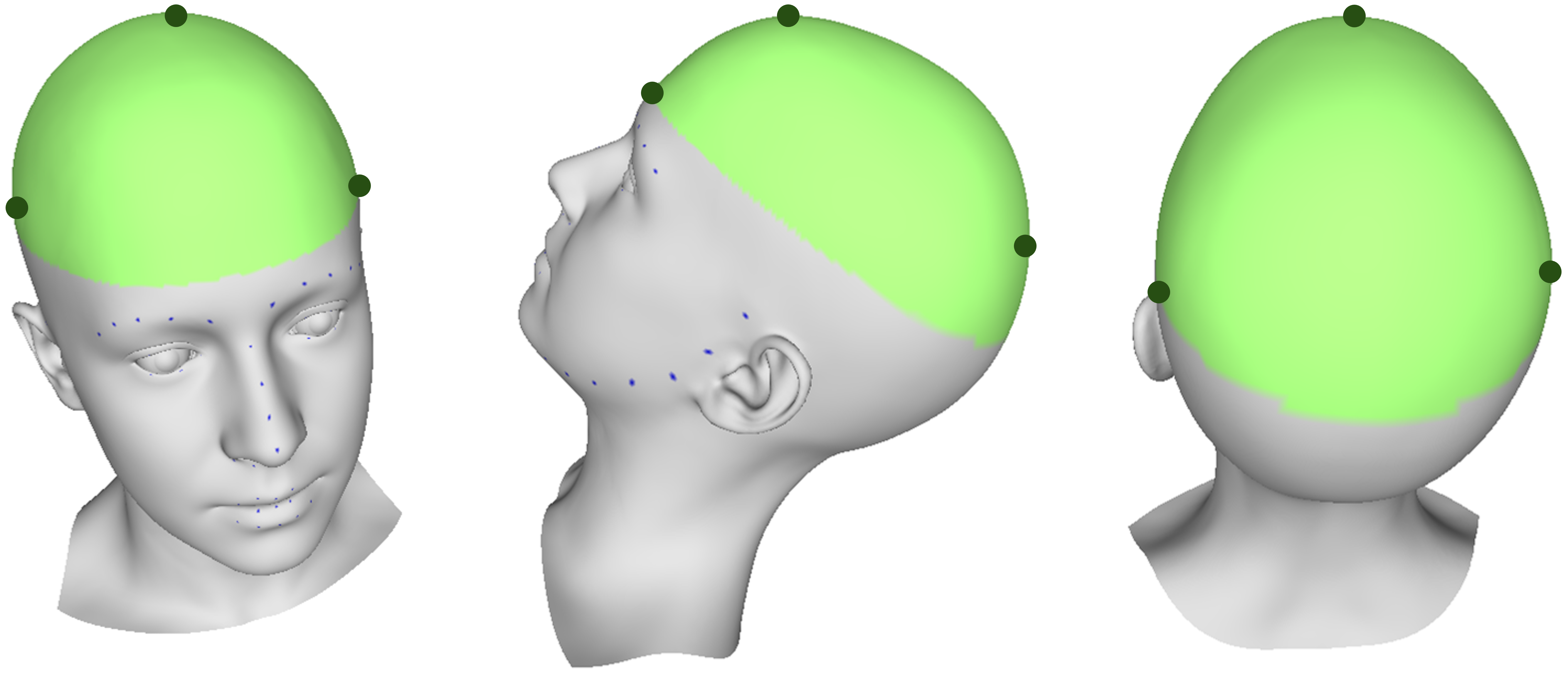} %
\caption{\textbf{Keypoints used for 3DMM fitting on the UHM standard model for sample views.} Keypoints include: i) facial landmarks (blue dots) corresponding to the facial landmarks extracted via \cite{bulat2017far}; and ii) view-dependent scalp features (green dots) marking the left-, right-, and top-most points on the upper head (green area).}
\label{fig:lmks}
\end{figure}
\begin{figure}[t]
\centering
\begin{tabular}{c@{\hskip 0.6in}c@{\hskip 0.5in}c@{\hskip 0.3in}r}
\textbf{a:} $I_n$ & \textbf{b:} $\mathcal{G}_n$ & \textbf{c:} $S(\alpha^{i=0})$  &  \textbf{d:} $S^{final}$ \\ 
\end{tabular}
\includegraphics[width=0.99\linewidth]{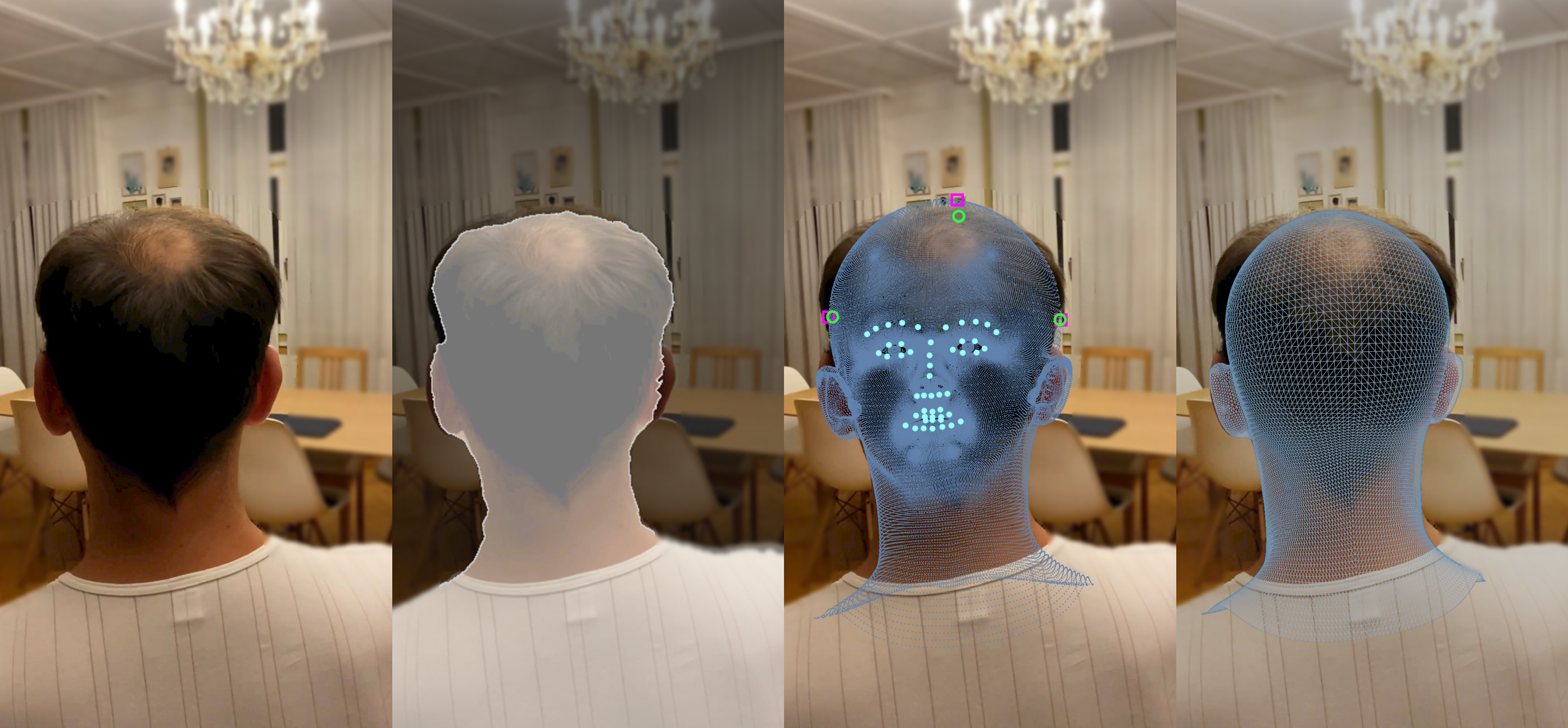} 
\caption{ \textbf{All-pose fitting.} a) Input image $I_n$; b) Silhouette $\mathcal{G}_n$ derived from $\mathcal{G}$; c) Projection of the first iteration $S(\alpha^{i=0})$ (\ie, $S^{mean}$) with facial landmarks (cyan dots) estimated from projection of the frontal fit and view-dependent scalp features (green dots) marking the left-, right-, and top-most points on the upper head (green area, Figure \ref{fig:lmks}) with the corresponding model scalp points (pink dots); d) Final fit $S^{final}$ projected on $I_n$.}
\label{fig:allPose}
\end{figure}

\textbf{All-pose fitting with adaptive scalp features (Stage 2).} 
Let $S^{top}$ denote the UHM region of the head above the ears as shown in Figure \ref{fig:lmks}. For each image $I_n$ in the $n$ sampled images from the complete pose range, we first define 3D scalp feature points on the UHM model as the left, right, and top-most point of $S_n(\alpha^i)$, obtained by using the retrieved current pose and the mesh $S(\alpha^i)$ fitted with the facial landmarks alone. Sample facial landmarks and scalp feature points are visualized for different views in Figure \ref{fig:lmks}, with blue and green dots respectively. Then, we find the corresponding 2D scalp feature points (x,y coordinates) on the image $I_n$ by selecting the corresponding (top-, left-, and right-most) points lying on the edge of the upper scalp silhouette $\mathcal{G}_n^{top}$
obtained by projecting the generic mesh $\mathcal{G}$ from Section \ref{sec:preprocess} onto each image $I_n$. 
We then rerun the align and fit optimization algorithm described in Equations \ref{eq:initAlign}-\ref{eq:multiImageFittingLin} where keypoints $X_k$ now include both scalp features and facial landmarks to obtain the final shape $S^{final}$ corresponding to shape parameters $\alpha_{final}$ that accounts for the scalp shape as well. Note that the scalp features detected on the UHM model are iteratively updated according to the alignment from the previous iteration. 
Note that, unlike \cite{hu2017efficient}, we use the perspective cameras from the SfM+MVS pipeline and we use nonlinear optimization to optimize the 6D pose of the UHM model in a multi-image setup rather than to estimate the camera in a single image setup. As shown in Figure \ref{fig:allPose}, the final fit now contours the head closely correcting the meshes obtained from Stage 1 (Figure \ref{fig:frontal}).

\section{Experiments}

\textbf{Dataset.} 
To test our algorithm, we created an initial dataset of $25$ subjects (ages 8-98) spanning multiple ethnicities. We asked the subjects to have a third person take 2 videos of their head at different times and locations with their smartphone devices, setting the resolution to a minimum of 1080p. 
Because videos were taken by different people at different locations, as can be noticed in the images throughout the paper, the guidelines were only loosely followed, providing us with a range of challenging backgrounds, illuminations, and head poses. 
All videos were converted to 1080p for processing.  

\textbf{Implementation details.}
For pre-processing, we uniformly sample $N=250$ frames per video.
Mesh filtering was fully automated using Meshroom's mesh filtering module, setting the hyperparameter value for the largest triangle to 30 and the number of smoothing iterations to 5, and keeping only the largest connected triangle group. 
Videos for which Meshroom failed to recover a mesh were discarded from further processing.
The number of frontal images $N_f$ used for the initial fit varied between $15-40$ depending on the video. We use half for determining $S^{front}$ and save the rest for evaluation purposes. 
For UHM fitting, we use all $600$ principal components of the original model. We run both stages of the fitting algorithm for $9$ iterations with regularization value $\lambda = 100$. 
The frames used for the all-pose fitting are obtained by sampling the images around the head every $15$ degrees (setting $\theta=0$ at the frontal/portrait frame used for the initial alignment) and restricting the absolute value of elevation to lie within $30$ degrees. This ensures that only the upper scalp portion 
is used to extract 
features (\vs accidentally including the neck area or ponytails).
The main algorithm (Stages 1-2) takes $\sim14$ minutes for each 1080p video on an Intel Xeon 32-core processor, with 128 GB RAM and 4 2080TI Nvidia GPUs. The entire pipeline, however, due to Meshroom pre-processing, final visualizations in Blender, and file IO can take up to 2 hours.

\textbf{Evaluations.}
Figure \ref{fig:Results1} shows sample qualitative results. For each person, we visualize the output fitted 3D heads $S^{final}$ mapped back on images from various viewing points (and head poses). $S^{final}$ closely contours the shape of the head for all subjects. Recall our focus to fit the head (face and scalp), so neck misalignment may occur since neck bending is not modeled by the UHM. 
We are unaware of a video dataset that goes around the complete head of each person (while also keeping the person still). 
Thus, in the absence of a 3D benchmark, we complement qualitative results with the following quantitative evaluations: 

\begin{figure*}[!htb]
 \includegraphics[width=0.98\linewidth]{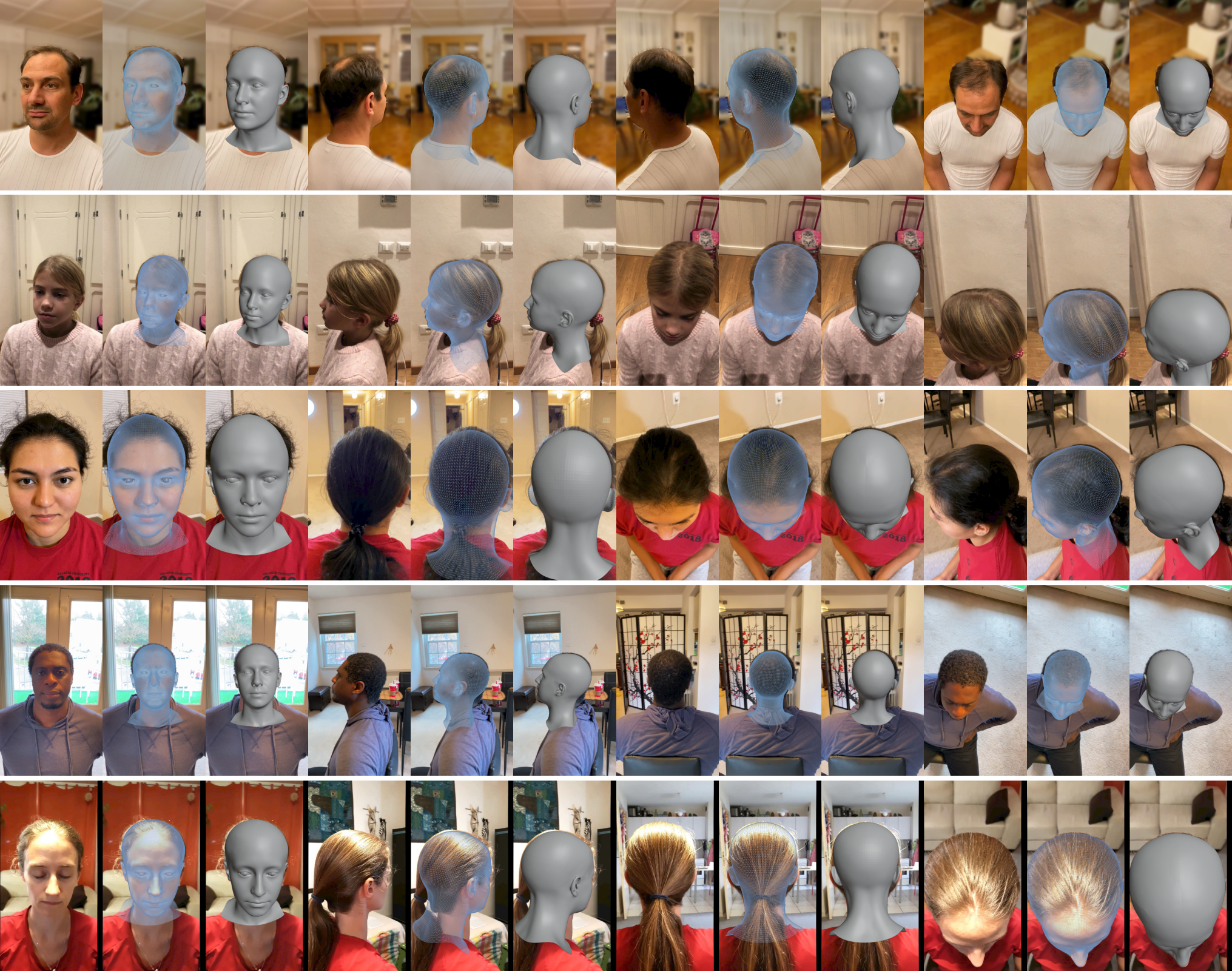}
  \caption{\textbf{Qualitative results.} For each person, we show the final projected mesh $S^{final}$ (wire-frame and solid visualizations) on sample frames. 
  Recall our focus to fit the head (face and scalp), so neck misalignment may occur since neck bending is not modeled by the UHM. 
  $S^{final}$ closely contour the shape of the head (in particular the upper scalp) for all subjects regardless of age or ethnicity.}
\label{fig:Results1}
\end{figure*}
%

\begin{table}[t!]
\centering
\begin{adjustbox}{max width=\linewidth}
\setlength{\tabcolsep}{2pt}
\begin{tabular}{l | c | c | c }
\toprule
        & $S^{mean}$ & $S^{front}$ & $S^{final}$  \\ 
\midrule
Height/Width & 4.2\% & 3.3\% & \bf{2.9\%} \\
Height/Length  &  9.9\% & 12.2\% &  \bf{5.7\%} \\
\bottomrule
\end{tabular}
\end{adjustbox}
\caption{\textbf{2D anthropometric GT evaluation.} Percent error between manually annotated head size ratios (Height/Width or Height/Length) and the average UHM shape $S^{mean}$, and the fits obtained at the two stages of our framework, $S^{front}$ and $S^{final}$.}
\label{tab:evalGTratios}
\end{table}

\textbf{i) 2D anthropometric ground-truth (GT) evaluation. } 
For head measurements, we created a simple anthropometric 2D GT by manually measuring width, length, and height in portrait ($\theta=0$) and lateral ($\theta=90$) images to compute head proportions (height over width and length over height, for the portrait and lateral views, respectively) and measure whether the reconstructed shapes adhered to the expected ratios. Measurements for each subject were taken twice for every dimension and averaged to minimize errors before comparing them with the automated ones. For each subject, we recorded the percent difference between GT ratios and the ratios automatically measured for the fits obtained at different steps of our method in Table \ref{tab:evalGTratios}.
$S^{front}$ shows poorer accuracy than $S^{mean}$, as it fits based on facial landmarks alone while relying on shape priors for the rest of the scalp. $S^{final}$ significantly improves after Stage 2 using our adaptive scalp features all-pose fitting. 

\begin{table}[t!]
\centering
\begin{adjustbox}{max width=\linewidth}
  \begin{tabular}{l@{\hskip 0.01in} |@{\hskip 0.01in}c@{\hskip 0.01in} |@{\hskip 0.01in}c@{\hskip 0.01in}|@{\hskip 0.01in}c@{\hskip 0.01in}}
\toprule
 & RMS lmks w/ jawlines & RMS lmks no jawline  & RMS jawline only \\ 
\midrule
\textbf{a:} Deepface 3D  & $13.7 \pm 6.7$ & $8.6\pm 3.4$ & $20.8\pm 10.6$\\
\textbf{b:} 3DFFA  & $16.9 \pm 7.3$ & $13.5 \pm5.6$  & $19.6\pm 8.8$\\
\textbf{c:} $S^{front}$ & $17.0 \pm 6.3$ & $14.7\pm 5.2$ & $ 20.3\pm 7.9 $\\
\textbf{d:} $S^{final}$ & $17.5 \pm 6.0$ & $ 15.5\pm 5.3$ & $20.7\pm 7.9$ \\
\textbf{e:} $S^{mean}$ & $19.4 \pm 6.4$ & $17.2\pm 6.5$ & $22.5 \pm 7.8$\\
\bottomrule
\end{tabular}
\end{adjustbox}
\caption{\textbf{Comparisons with state-of-the-art 3D face reconstruction methods.} We evaluate the root mean squared (RMS) error (in pixels) between 2D facial landmarks detected on withheld frontal views (\ie, not used for any fitting) projected fit obtained by:  Deepface 3D \cite{deng2019accurate}(a); 3DDFA \cite{3ddfa_cleardusk, guo2020towards, zhu2017face} (b); our algorithm after Stage 1, \ie the frontal mesh $S^{front}$ (c); our final output mesh $S^{final}$ (d); and, for baseline comparison, the average UHM shape $S^{mean}$ (e).}
\label{tab:evalComparison}
\end{table}
%
\begin{table}[t!]
\centering
\begin{adjustbox}{max width=\linewidth}
\setlength{\tabcolsep}{2pt}
\begin{tabular}{l | c | c | c}
\toprule
 Chamfer distance (mm) & $S^{mean}$ & $S^{front}$ &  $S^{final}$ \\ 
\midrule $\mathcal{G}$ &  6.72 &  8.32   & 4.80\\
\bottomrule
\end{tabular}
\end{adjustbox}
\caption{\textbf{Fitted scalp \vs $\mathcal{G}$.}
Chamfer distance measured between $\mathcal{G}$ and the fitted heads for the upper scalp $S^{top}$, using a reference head model width of 16 cm. 
}
\label{tab:Chamfer}
\vspace{-5pt}
\end{table}

%
%
\begin{table}[t]
\centering
\begin{adjustbox}{max width=\linewidth}
\setlength{\tabcolsep}{2pt}
\begin{tabular}{l | c | c | c}
\toprule
Chamfer distance (mm)   & $S^{mean}$ & $S^{front}$ &  $S^{final}$ \\ 
\midrule 
\bf{a:}  $\mathcal{G}_{lidar}$ &  4.64 & 5.60   & 3.20\\
\midrule 
\bf{b:} $\mathcal{G}$ & 5.12 & 6.56   & 4.00\\
\bottomrule
\end{tabular}
\end{adjustbox}
\caption{\textbf{Fitted scalp \vs LiDAR reconstruction.} 
Chamfer distance for:
\textbf{a:} $\mathcal{G}_{Lidar}$, the 3D reconstruction from an Intel RealSense L515 LiDAR camera video; \textbf{b:} $\mathcal{G}$ for the same 4 subjects. 
}
\label{tab:ChamferLidar}
\end{table}

\textbf{ii) Comparison with state-of-the-art (SOTA) 3D face reconstruction methods on withheld views.} 
We compare our final fits with two SOTA 3DMM fitting models: 3DDFA \cite{3ddfa_cleardusk, guo2020towards, zhu2017face} and Deepface 3D reconstruction \cite{deng2019accurate} on a total of $\sim800$ images from the 25 subjects used for training and testing.
The goal of this evaluation is to show how the proposed method compares with the SOTA and not to outperform these methods. SOTA methods, while better for the face, do not extend beyond the forehead and thus cannot be used for the scalp.
Table \ref{tab:evalComparison} shows the root mean squared (RMS) 2D Euclidean distance (in pixels) between the projected landmarks and the ones detected by the facial detector used in Section \ref{sec:preprocess}.
Both SOTA methods were evaluated on single image fitting (\vs our multi-image one) which generally leads to better reprojection error. RMS for $S^{final}$ is computed using landmarks projected by $S^{front}$ on new images sampled from the complete pose ranges, not the original images that were used in the fitting. 
Additionally, in our videos, though we recommended a neutral expression,  people closed their eyes or were smiling. Both Deepface and 3DDFA handle expressions leading to better metrics for the face. We find comparable results for jawline landmarks on their own, since the expressions observed in the videos affect the jawlines to a much lower degree.

\begin{figure}[t]
\centering
\includegraphics[width=0.97\linewidth]{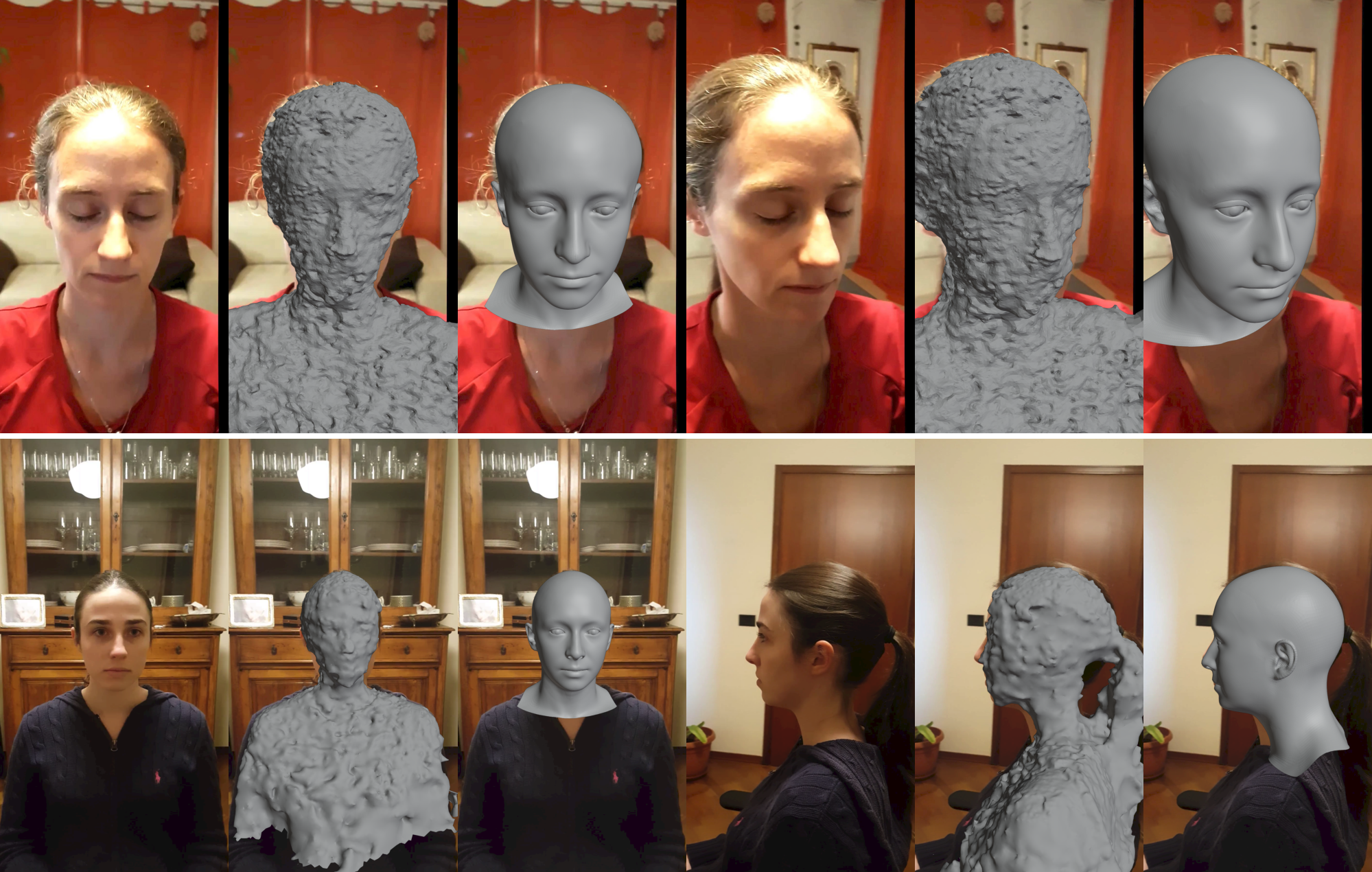} 
\caption{\textbf{Robustness of two-stage fitting.} For two sample subjects (displaying sample input frame, $\mathcal{G}$, and $S^{final}$), we show robustness of the method even in the presence of very jagged or incorrect dense reconstructions $\mathcal{G}$.}
\label{fig:benefitsvsSfM}
\end{figure}

\textbf{iii) 3D comparison of fitted scalp \vs $\mathcal{G}$.}
In Table \ref{tab:Chamfer}, we compute the Chamfer distance to compare the fitted scalp mesh with the original dense reconstruction $\mathcal{G}$ for the scalp portion of the head $S^{top}$. We exclude the facial portion because the high degree of curvature and edges contained on the faces can lead to extremely jagged $\mathcal{G}$ meshes. For each vertex belonging to $S^{top}$, aligned in the SfM canonical space, we measure the distance between this vertex and its corresponding closest vertex in $\mathcal{G}$. We then compute the mean distance between the UHM scalp vertices and $\mathcal{G}$'s vertices and divide it by the width of the head of the UHM model ($T_{opt}T_{sim}S^{final}$). 
We qualitatively show the benefit of our fitting pipeline instead of directly using the dense reconstruction in Figure \ref{fig:benefitsvsSfM}. Subpar illumination conditions or image quality can lead to poor 3D reconstructions. 
In our experiments, we find that the dense reconstruction from the SfM+MVS pipeline can be unreliable for regions of high curvature such as jawlines or nose, while still providing reliable camera information. 
Our approach is robust to such scenarios and further justifies use of nonlinear optimization in the two-stage fitting algorithm.

\textbf{iv) 3D comparison with 3D LiDAR reconstruction.} We include 3D GT evaluation on a small subset of our original subjects by capturing the shape of the scalp with a tight head scarf to minimize further any remaining hair volume. Using an Intel RealSense L515 LiDAR camera, we took videos of four subjects, who were physically located near our lab. We generated the complete 3D LiDAR mesh using the Open3D library~\cite{Zhou2018}. We then measured the Chamfer distance between our output and the GT LiDAR mesh $\mathcal{G}_{lidar}$ (scale-normalized to the default UHM model). As shown in Table \ref{tab:ChamferLidar}, we find similar trends with Table \ref{tab:Chamfer} and our final meshes $S^{final}$ are closer to the LiDAR scans than to $\mathcal{G}$.

\begin{figure*}[!htb]
\includegraphics[width=0.98\linewidth]{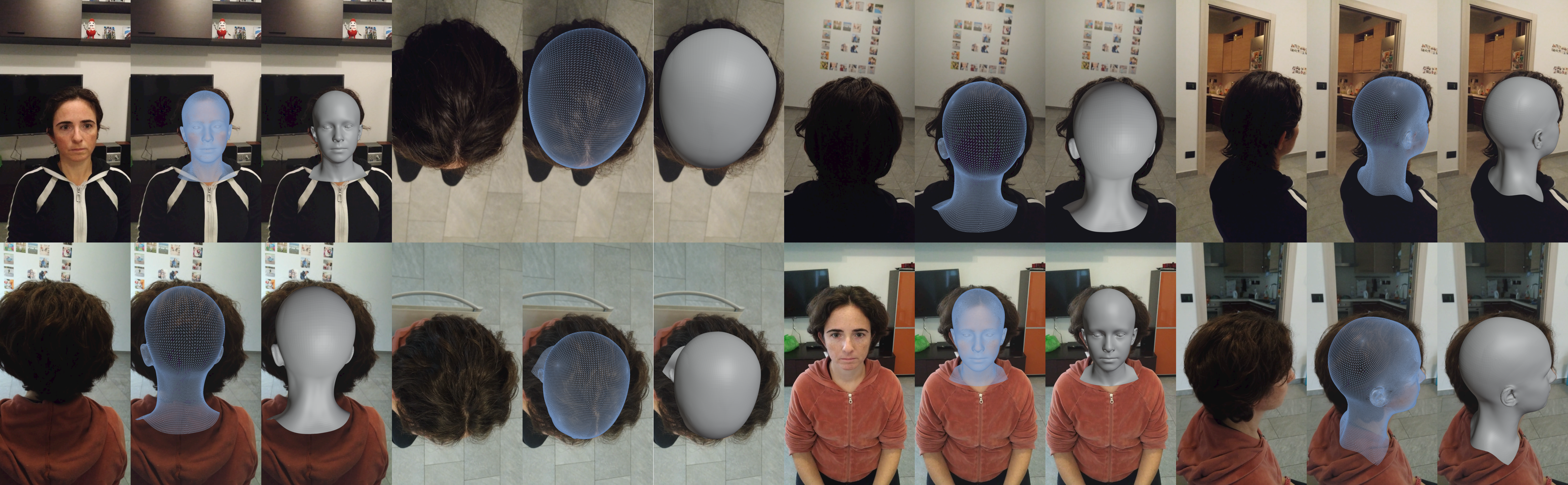}\\
\includegraphics[width=0.98\linewidth]{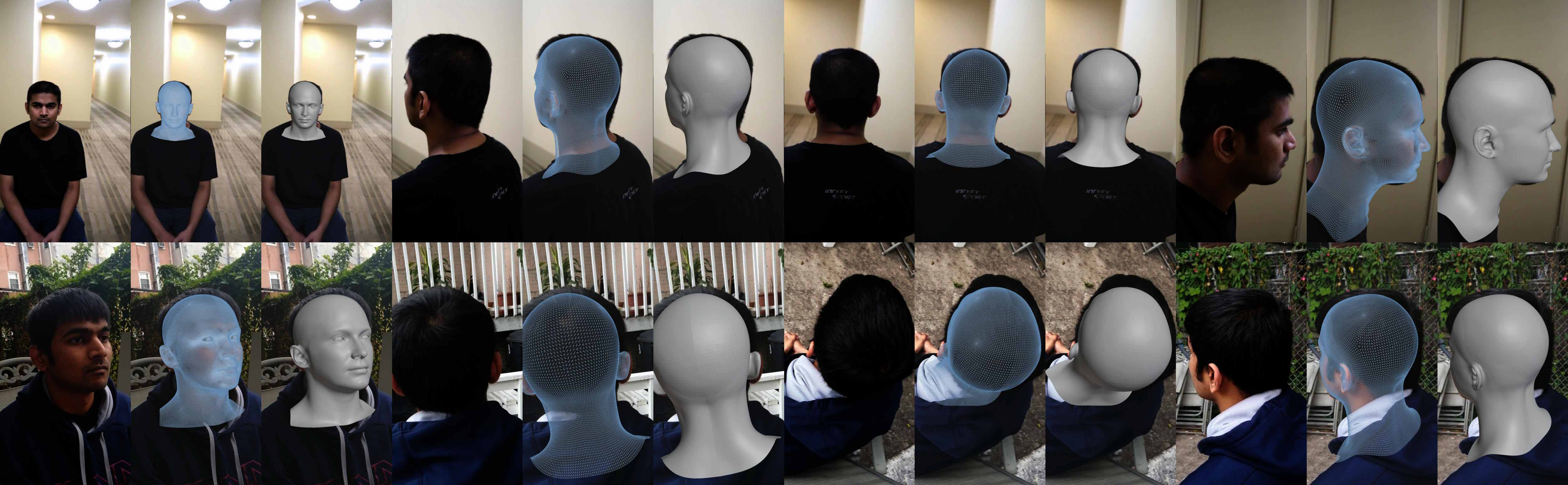}
  \caption{\textbf{Qualitative results on new videos with unconstrained hairstyles.} For each subject, we show final meshes on two videos: in the first row, the video used to create the 3D head; in the second row, the final fitted mesh on a new video with larger hair volume and/or unrestricted hairstyle.}
\label{fig:UHMswapping}
\end{figure*}

\textbf{Fitting reproducibility.} Finally, we show the full potential of our method by evaluating mesh consistency and demonstrating how, once a 3D mesh for a person's head is obtained, it can be used for other videos of the same person, in particular the ones in which the required assumptions to recover the scalp shape no longer need to hold. This applies to both different hairstyles 
and larger hair volumes. 
To measure reconstruction consistency between two meshes from different videos of the same person, the average vertex displacement error over all vertices for the entire head and for the separate face and scalp regions yielded $1.30\%$,
$2.37\%$, and $1.09\%$ errors on the overall head, face, and scalp areas, respectively.
In Figure \ref{fig:UHMswapping}, we qualitatively show that, once we have obtained the shape of an individual, we need not determine the shape again and can simply determine $T_{opt}$ and $T_{sim}$ for $S(\alpha)$ using just a few frontal images to obtain accurate mapping for all poses.

\section{Discussion and Conclusions}
We present a new framework to recover a person's head geometry from a video taken with a single handheld camera.
Results on $25$ subjects show that the method works on a variety of videos. In all experiments, results quantitatively indicate that scalp features included in the two-stage algorithm lead to better fitting than the 3DMM prior alone.
The current approach suffers from a few limitations. First, the implementation is computationally expensive as each video can take up to 2 hours using high-end hardware. Second, requiring the head to remain still while moving the camera does not allow a person to take the video on their own. Third, it does not include ear keypoints since we did not find a reliable automated method to detect features for the ear.
Finally, it relies on the 3D generic reconstruction as well as facial landmark detector as initial inputs. If one or both happen to fail, then the video has to be retaken. 

The proposed head fitting method yields a high degree of consistency between fitted heads from different videos of the same person.
Because the alignment with the 3D surface mesh can be done starting with any 3D head model, the pipeline presented can also be used to align subsequent videos by using a 3D head fitted from a different video (of the same person). This is particularly useful for certain clinical applications such as mapping hair loss information since specific types of hair loss require having the hair parted in different ways to expose the hair loss.  
Finally, as a first step towards creating an end-to-end system to solve the all-pose head fitting problem, this approach can be used to generate training data for deep learning models to enable fitting morphable models across all pose ranges.
The fitted 3D heads will be used, in future work, for mapping texture information onto the subject's head surface for tracking clinically valuable information. The consistent mesh topology inherent in the UHM model will simplify tracking as it implicitly provides point-to-point correspondences for the entire head surface, setting the path for many clinical applications.

\noindent
\textbf{Acknowledgements. }
This research was in part funded by a research grant by the Edwin and Fannie Gray Hall Center for Human Appearance, at the University of Pennsylvania, awarded to Drs. Bernardis and Daniilidis. 

{\small
\bibliographystyle{ieee_fullname}
\bibliography{arxiv.bib}

\begin{thebibliography}{10}\itemsep=-1pt

\bibitem{Agrawal_2020_WACV}
Shubham Agrawal, Anuj Pahuja, and Simon Lucey.
\newblock High accuracy face geometry capture using a smartphone video.
\newblock In {\em IEEE Workshop on Applic of Comput Vis}, March 2020.

\bibitem{alexander2013digital}
Oleg Alexander, Graham Fyffe, Jay Busch, Xueming Yu, Ryosuke Ichikari, Andrew
  Jones, Paul Debevec, Jorge Jimenez, Etienne Danvoye, Bernardo Antionazzi,
  et~al.
\newblock Digital ira: Creating a real-time photoreal digital actor.
\newblock In {\em ACM SIGGRAPH Posters}, pages 1--1. 2013.

\bibitem{Meshroom}
AliceVision.
\newblock {Meshroom: A 3D reconstruction software}.
\newblock \url{https://github.com/alicevision/meshroom}, 2018.

\bibitem{Alldieck2018}
Thiemo Alldieck, Marcus Magnor, Weipeng Xu, Christian Theobalt, and Gerard
  Pons-Moll.
\newblock Detailed human avatars from monocular video.
\newblock In {\em Int Conf on 3D Vision}, pages 98--109. IEEE, 2018.

\bibitem{badger20203d}
Marc Badger, Yufu Wang, Adarsh Modh, Ammon Perkes, Nikos Kolotouros, Bernd~G
  Pfrommer, Marc~F Schmidt, and Kostas Daniilidis.
\newblock 3d bird reconstruction: a dataset, model, and shape recovery from a
  single view.
\newblock {\em arXiv:2008.06133}, 2020.

\bibitem{Beeler2010}
Thabo Beeler, Bernd Bickel, Paul Beardsley, Bob Sumner, and Markus Gross.
\newblock High-quality single-shot capture of facial geometry.
\newblock {\em ACM Trans on Graph}, 29(4), July 2010.

\bibitem{blanz1999morphable}
Volker Blanz and Thomas Vetter.
\newblock A morphable model for the synthesis of 3d faces.
\newblock In {\em Conf on Comput Graph and Interactive Techniques}, pages
  187--194, 1999.

\bibitem{Booth2018_IJCV}
James Booth, Anastasios Roussos, Allan Ponniah, David Dunaway, and Stefanos
  Zafeiriou.
\newblock Large scale 3d morphable models.
\newblock {\em Int J Comput Vision}, 126(2–4):233–254, Apr. 2018.

\bibitem{bulat2017far}
Adrian Bulat and Georgios Tzimiropoulos.
\newblock How far are we from solving the 2d \& 3d face alignment problem? (and
  a dataset of 230,000 3d facial landmarks).
\newblock In {\em IEEE Int Conf on Comput Vis}, 2017.

\bibitem{cao2018stabilized}
Chen Cao, Menglei Chai, Oliver Woodford, and Linjie Luo.
\newblock Stabilized real-time face tracking via a learned dynamic rigidity
  prior.
\newblock {\em ACM Trans on Graph}, 37(6), 2018.

\bibitem{cao2013facewarehouse}
Chen Cao, Yanlin Weng, Shun Zhou, Yiying Tong, and Kun Zhou.
\newblock Facewarehouse: A 3d facial expression database for visual computing.
\newblock {\em IEEE J VCG}, 20(3):413--425, 2013.

\bibitem{Cao2018}
X. {Cao}, Z. {Chen}, A. {Chen}, X. {Chen}, S. {Li}, and J. {Yu}.
\newblock Sparse photometric 3d face reconstruction guided by morphable models.
\newblock In {\em IEEE Int Conf on Comput Vis and Patt Recognit}, pages
  4635--4644, 2018.

\bibitem{Dai2019}
Hang Dai, Nick Pears, William Smith, and Christian Duncan.
\newblock Statistical modeling of craniofacial shape and texture.
\newblock {\em Int J Comput Vision}, 128(2):547--571, Nov 2019.

\bibitem{deng2019accurate}
Yu Deng, Jiaolong Yang, Sicheng Xu, Dong Chen, Yunde Jia, and Xin Tong.
\newblock Accurate 3d face reconstruction with weakly-supervised learning: From
  single image to image set.
\newblock In {\em IEEE Int Conf on Comput Vis and Patt Recognit { Workshops}},
  2019.

\bibitem{furukawa15}
Yasutaka Furukawa and Carlos Hern{\'a}ndez.
\newblock Multi-view stereo: A tutorial.
\newblock {\em Foundations and Trends in Computer Graphics and Vision}, 2015.

\bibitem{debevec_facialScans}
Graham Fyffe, Andrew Jones, Oleg Alexander, Ryosuke Ichikari, and Paul Debevec.
\newblock Driving high-resolution facial scans with video performance capture.
\newblock {\em ACM Trans on Graph}, 34(1), Dec. 2015.

\bibitem{fyffe_multi-view_2017}
G. Fyffe, K. Nagano, L. Huynh, S. Saito, J. Busch, A. Jones, H. Li, and P.
  Debevec.
\newblock Multi-{View} {Stereo} on {Consistent} {Face} {Topology}.
\newblock {\em Comput Graph Forum}, 36(2):295--309, May 2017.

\bibitem{goel2020shape}
Shubham Goel, Angjoo Kanazawa, and Jitendra Malik.
\newblock Shape and viewpoint without keypoints.
\newblock {\em arXiv:2007.10982}, 2020.

\bibitem{3ddfa_cleardusk}
Jianzhu Guo, Xiangyu Zhu, and Zhen Lei.
\newblock {3DDFA}.
\newblock \url{https://github.com/cleardusk/3DDFA}, 2018.

\bibitem{guo2020towards}
Jianzhu Guo, Xiangyu Zhu, Yang Yang, Fan Yang, Zhen Lei, and Stan~Z Li.
\newblock Towards fast, accurate and stable 3d dense face alignment.
\newblock In {\em IEEE European Conf on Comput Vis}, 2020.

\bibitem{Hartley:2003:MVG:861369}
Richard Hartley and Andrew Zisserman.
\newblock {\em Multiple View Geometry in Computer Vision}.
\newblock Cambridge University Press, New York, NY, USA, 2 edition, 2003.

\bibitem{hu2017efficient}
Guosheng Hu, Fei Yan, Josef Kittler, William Christmas, Chi~Ho Chan, Zhenhua
  Feng, and Patrik Huber.
\newblock Efficient 3d morphable face model fitting.
\newblock {\em Lect Notes Comput Sc}, 67:366--379, 2017.

\bibitem{hu2017avatar}
Liwen Hu, Shunsuke Saito, Lingyu Wei, Koki Nagano, Jaewoo Seo, Jens Fursund,
  Iman Sadeghi, Carrie Sun, Yen-Chun Chen, and Hao Li.
\newblock Avatar digitization from a single image for real-time rendering.
\newblock {\em ACM Trans on Graph}, 36(6):1--14, 2017.

\bibitem{huber2016multiresolution}
Patrik Huber, Guosheng Hu, Rafael Tena, Pouria Mortazavian, P Koppen, William~J
  Christmas, Matthias Ratsch, and Josef Kittler.
\newblock A multiresolution 3d morphable face model and fitting framework.
\newblock In {\em International Joint Conference on Computer Vision, Imaging
  and Computer Graphics Theory and Applications}, 2016.

\bibitem{Jancosek2011}
Michal Jancosek and Tomas Pajdla.
\newblock Multi-view reconstruction preserving weakly-supported surfaces.
\newblock In {\em IEEE Int Conf on Comput Vis and Patt Recognit}, jun 2011.

\bibitem{kanazawa2018learning}
Angjoo Kanazawa, Shubham Tulsiani, Alexei~A. Efros, and Jitendra Malik.
\newblock Learning category-specific mesh reconstruction from image
  collections.
\newblock {\em arXiv:1803.07549}, 2018.

\bibitem{kim2018deep}
Hyeongwoo Kim, Pablo Garrido, Ayush Tewari, Weipeng Xu, Justus Thies, Matthias
  Niessner, Patrick P{\'e}rez, Christian Richardt, Michael Zollh{\"o}fer, and
  Christian Theobalt.
\newblock Deep video portraits.
\newblock {\em ACM Trans on Graph}, 37(4):1--14, 2018.

\bibitem{dlib09}
Davis~E. King.
\newblock Dlib-ml: A machine learning toolkit.
\newblock {\em J Mach Learn Res}, 10:1755--1758, 2009.

\bibitem{lee1995realistic}
Yuencheng Lee, Demetri Terzopoulos, and Keith Waters.
\newblock Realistic modeling for facial animation.
\newblock In {\em Conf on Comput Graph and Interactive Techniques}, pages
  55--62, 1995.

\bibitem{FLAME:SiggraphAsia2017}
Tianye Li, Timo Bolkart, Michael.~J. Black, Hao Li, and Javier Romero.
\newblock Learning a model of facial shape and expression from {4D} scans.
\newblock {\em ACM Trans on Graph}, 2017.

\bibitem{3DHair_from_video_Ira}
Shu Liang, Xiufeng Huang, Xianyu Meng, Kunyao Chen, Linda~G. Shapiro, and Ira
  Kemelmacher-Shlizerman.
\newblock Video to fully automatic 3d hair model.
\newblock {\em ACM Trans on Graph}, 37(6):206, Dec. 2018.

\bibitem{SMPL:2015}
Matthew Loper, Naureen Mahmood, Javier Romero, Gerard Pons-Moll, and Michael~J.
  Black.
\newblock {SMPL}: A skinned multi-person linear model.
\newblock {\em ACM Trans on Graph}, 34(6):248:1--248:16, Oct. 2015.

\bibitem{more1978levenberg}
Jorge~J Mor{\'e}.
\newblock The levenberg-marquardt algorithm: implementation and theory.
\newblock In {\em Numerical analysis}, pages 105--116. Springer, 1978.

\bibitem{Moulon2012}
Pierre Moulon, Pascal Monasse, and Renaud Marlet.
\newblock Adaptive structure from motion with a~contrario model estimation.
\newblock In {\em Asian Conf on Comput Vis}, pages 257--270. Springer Berlin
  Heidelberg, 2012.

\bibitem{Nam2019_hair3D}
Giljoo Nam, Chenglei Wu, Min~H. Kim, and Yaser Sheikh.
\newblock Strand-accurate multi-view hair capture.
\newblock In {\em IEEE Int Conf on Comput Vis and Patt Recognit}, June 2019.

\bibitem{bfm09}
P. Paysan, R. Knothe, B. Amberg, S. Romdhani, and T. Vetter.
\newblock A 3d face model for pose and illumination invariant face recognition.
\newblock In {\em IEEE Int Conf on Advanced Video and Signal based Surveillance
  (AVSS) for Security, Safety and Monitoring in Smart Environments}, 2009.

\bibitem{ploumpis2020towards}
Stylianos Ploumpis, Evangelos Ververas, Eimear O'Sullivan, Stylianos
  Moschoglou, Haoyang Wang, Nick Pears, William Smith, Baris Gecer, and
  Stefanos~P Zafeiriou.
\newblock Towards a complete 3d morphable model of the human head.
\newblock {\em IEEE Trans on Patt Anal and Mach Inte}, 2020.

\bibitem{schonberger2016structure}
Johannes~L Sch\"{o}nberger and Jan-Michael Frahm.
\newblock Structure-from-motion revisited.
\newblock In {\em IEEE Int Conf on Comput Vis and Patt Recognit}, pages
  4104--4113, 2016.

\bibitem{schoenberger2016mvs}
Johannes~Lutz Sch\"{o}nberger, Enliang Zheng, Marc Pollefeys, and Jan-Michael
  Frahm.
\newblock Pixelwise view selection for unstructured multi-view stereo.
\newblock In {\em IEEE European Conf on Comput Vis}, October 2016.

\bibitem{tran2018nonlinear}
Luan Tran and Xiaoming Liu.
\newblock Nonlinear 3d face morphable model.
\newblock In {\em IEEE Int Conf on Comput Vis and Patt Recognit}, pages
  7346--7355, 2018.

\bibitem{tuan2017regressing}
Anh Tuan~Tran, Tal Hassner, Iacopo Masi, and G{\'e}rard Medioni.
\newblock Regressing robust and discriminative 3d morphable models with a very
  deep neural network.
\newblock In {\em IEEE Int Conf on Comput Vis and Patt Recognit}, pages
  5163--5172, 2017.

\bibitem{umeyama1991least}
Shinji Umeyama.
\newblock Least-squares estimation of transformation parameters between two
  point patterns.
\newblock {\em IEEE Trans on Patt Anal and Mach Inte}, 4:376--380, 1991.

\bibitem{Zhou2018}
Qian-Yi Zhou, Jaesik Park, and Vladlen Koltun.
\newblock {Open3D}: {A} modern library for {3D} data processing.
\newblock {\em arXiv:1801.09847}, 2018.

\bibitem{Zhou_2018_ECCV}
Yi Zhou, Liwen Hu, Jun Xing, Weikai Chen, Han-Wei Kung, Xin Tong, and Hao Li.
\newblock Hairnet: Single-view hair reconstruction using convolutional neural
  networks.
\newblock In {\em IEEE European Conf on Comput Vis}, pages 235--251, September
  2018.

\bibitem{zhu2017face}
Xiangyu Zhu, Xiaoming Liu, Zhen Lei, and Stan~Z Li.
\newblock Face alignment in full pose range: A 3d total solution.
\newblock {\em IEEE Trans on Patt Anal and Mach Inte}, 2017.

\bibitem{zuffi2019three}
Silvia Zuffi, Angjoo Kanazawa, Tanya Berger-Wolf, and Michael Black.
\newblock Three-d safari: Learning to estimate zebra pose, shape, and texture
  from images in the wild.
\newblock In {\em IEEE Int Conf on Comput Vis}, pages 5358--5367, 2019.

\end{thebibliography}
}

\clearpage

\end{document}